\documentclass{article}
\usepackage{ijcai19}

\usepackage{times}
\usepackage{soul}
\usepackage{url}
\usepackage[hidelinks]{hyperref}
\usepackage[utf8]{inputenc}
\usepackage[small]{caption}
\usepackage{graphicx}
\usepackage{amsmath}
\usepackage{booktabs}
\usepackage[ruled,linesnumbered,noline]{algorithm2e}
\usepackage{paralist}
\usepackage{enumitem}
\usepackage{comment}
\usepackage{multirow}
\usepackage{tabularx}
\usepackage{makecell}
\usepackage[skip=4pt]{caption}

\newcolumntype{P}[1]{>{\centering\arraybackslash}p{#1}}

\DeclareMathOperator*{\argmax}{arg\,max}
\DeclareMathOperator*{\argmin}{arg\,min}
\title{DARec: Deep Domain Adaptation for Cross-Domain Recommendation via Transferring Rating Patterns}

\author{
Feng Yuan$^1$
\and
Lina Yao$^1$\And
Boualem Benatallah$^1$
\affiliations
$^1$University of New South Wales\\
}

\begin{document}

\maketitle

\begin{abstract}

Cross-domain recommendation has long been one of the major topics in recommender systems.Recently, various deep models have been proposed to transfer the learned knowledge across domains, but most of them focus on extracting abstract transferable features from auxilliary contents, e.g., images and review texts, and the patterns in the rating matrix itself is rarely touched. In this work, inspired by the concept of domain adaptation, we proposed a deep domain adaptation model (DARec) that is capable of extracting and transferring patterns from rating matrices {\em only} without relying on any auxillary information. We empirically demonstrate on public datasets that our method achieves the best performance among several state-of-the-art alternative cross-domain recommendation models.

\end{abstract}

\section{Introduction}
Recommender systems have long been beset by a host of well-known issues, e.g., the cold-start and sparsity problems. As one solution, cross-domain recommendation (CDR) leverages the information from several different source systems or domains to build a better recommendation model for the target domain. In real-world datasets, although the number of users and items are normally very large with limited available feedbacks, the users/items can be normally categorized into groups (domains) according to certain criteria, making CDR a promising and practical technique. During the past few decades, CDR has been studied from a number of perspectives in different research areas, e.g., various domain definitions, recommendation scenarios and recommendation tasks \cite{khan2017cross}. For instance, recommendation scenarios can be divided into four classes: \begin{inparaenum} \item No User–No Item overlap (NU-NI); \item User–No Item overlap (U-NI); \item No User–Item overlap (NU-I); \item User–Item overlap (U-I). \end{inparaenum} In this work, we focus on single-domain recommendation task under the scenario where users are fully aligned. In other words, we model rating patterns in both source and target domains, and enhance the recommendations in the target domain.

Under our research settings, a number of approaches have been proposed. Some are based on clustering \cite{li2009RMGM}, and others use variations of matrix factorization (MF) \cite{singh2008CMF}. A few recent works apply deep learning (DL) techniques to perform the knowledge transfer \cite{elkahky2015MVDNN,lian2017CCCFNet}. Most clustering and MF-based methods fail to model the nonlinear patterns in the ratings and normally require a dense rating matrix for the source domain. DL methods show superior performance to the above ones, but most of them learn and transfer knowledge from auxilliary contents (e.g., item features). Directly extracting and transferring patterns from the sparse rating matrix are rarely studied. Since DL have been reported to have state-of-the-art performance in both rating prediction and top-N recommendation tasks by only using rating matrices \cite{he2017neural,he2018convmf}, we propose to leverage deep neural networks (DNN) to learn the transferable rating patterns.

Domain adaptation (DA) is a technique that allows knowledge from a source domain to be transferred to a different but related target domain. DA is widely employed in semi-supervised learning, where the target domain are not labelled or have only a few labels. DA aims to learn a classifier or predictor in presence of a distribution shift between the source and target domain, and has been shown to have state-of-the-art performance in various computer vision and natural language processing tasks \cite{bousmalis2016DSN}.

It is true that typical recommendation problems do not exactly match the application scenarios of DA. However, as a fundamental question in cross-domain recommendations, how to model the data-dependent effect of the ratings when transferring the data-independent knowledge needs to be addressed. Since the source and target data may be with different distributions, in order to extract the shared rating patterns from the two domains, we are inspired by the domain adversarial neural networks (DANN) \cite{ganin2015da} and propose a deep Domain Adaptation Recommendation (DARec) model that is composed of a rating pattern extractor, a domain classifier and a predictor for rating estimation tasks. It can automatically learn abstract representations of shared patterns and transfer them between two domains via DNN. Furthermore, we apply a weighted loss function to balance the significance of the source and target domains to control the transfer direction. Due to the extreme sparsity in the original rating matrices, we first apply AutoRec \cite{sedhain2015autorec} to learn a dense embedding for each user to represent the user's preferences, after which the embeddings are fed into the following DNN as inputs. We show that using DARec, the rating estimation accuracy in the target domain can be largely improved. We further compare our approach with several state-of-the-art CDR methods on public datasets, showing that DARec achieves best recommendation results. In summary, our major contributions are as follows:
\begin{itemize}
    \item We propose a deep domain adaptation cross-domain recommendation model (DARec) to extract and transfer abstract rating patterns, using only the information from  rating matrices. Utilizing a domian classifier, shared rating patterns of the same user in different domains are learned and transferred via adversarial training, leading to U-DARec model, while distinctive rating patterns of unrelated items in different domains are seperated, resulting in the I-DARec model;
    \item Using serveral public datasets, we empirically investigate the capability of our approach to improve on the recommendation performance in single-domain, which confirms the effectiveness of DARec to transfer knowledge between domains. We compare our model with some state-of-the-art single-domain rating prediction methods and demonstrate that benefiting from the transferred information from the source domain, DARec is superior to all the selected baselines;
    \item We select several alternative cross-domain recommendation methods that can fit in our research settings, including MF-based models and recently-proposed DL-based ones. On the public datasets, we demonstrate that due to DL, DARec is superior to all the selected MF models, and I-DARec exceeds the performance of state-of-the-art DL-based methods.
\end{itemize}

\begin{figure*}[!ht]
    \centering
    \includegraphics[width=\textwidth,height=6.5cm]{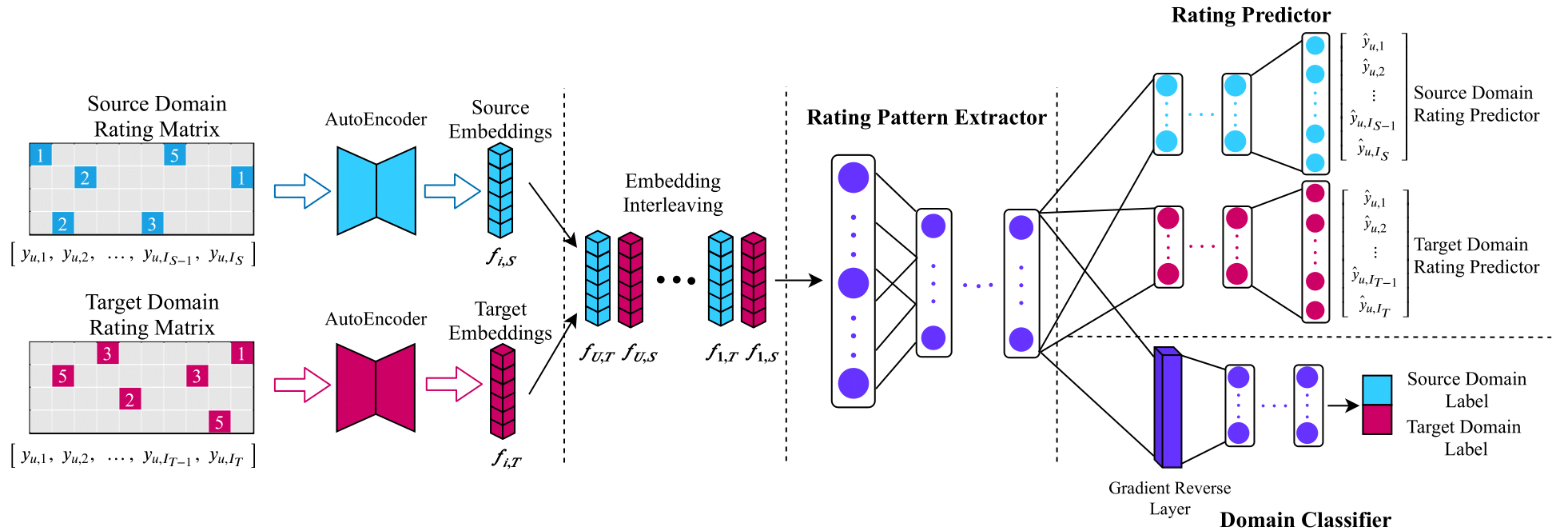}
    \caption{The Proposed Structure of Deep Domain Adaptation-Based Cross-Domain Recommendation}
    \label{fig:DARec}
    \vspace{-4mm}
\end{figure*}

\section{Related Work}
Generally, CDR can be categorized as collaborative filtering-based CDR (CFCDR) methods and content-based CDR (CBCDR) methods.Early CFCDR approaches use neighborhood-based solutions \cite{berkovsky2007cross}, but these methods are quickly surpassed by MF-based solutions. Collective Matrix Factorization (CMF) \cite{singh2008CMF} couples target rating matrix and all auxiliary matrices on \textit{User} dimension to share the user factor matrix across all domains. Cross-domain Triadic Factorization (CDTF) \cite{hu2013personalized} learns the triadic factors for user, item and domain through tensor factorization, which can better capture the interactions between domain-specific user factors and item factors. Cluster-level Matrix Factorization \cite{mirbakhsh2015improving} employs SVD and K-means clustering algorithm to build a cross-domain cluster-level coarse matrix that captures shared patterns between the cluster of users and the cluster of items among multiple domains. Most of the MF-based CFCD models have two major shortcomes: \begin{inparaenum} \item most of them assume the auxiliary data is relative dense for all users or items; \item user/item overlaps are normally required \cite{cremonesi2011cross}; \item the linear MF methods are not powerful enough to extract nonlinear patterns \end{inparaenum}.

To overcome such limitations, CBCDR techniques are proposed to establish domain links by leveraging the auxiliary information about users/items without requiring user/item overlaps or relative dense rating matrices. Moreover, DL is ready to be applied on such auxiliary data to learn deep features that can better represent users/items. \cite{elkahky2015MVDNN} is a multi-view framework that extends the deep structured semantic model which maps high-dimensional features into low-dimensional dense features in a joint semantic space, to merge more than two views of the auxiliary data. \cite{lian2017CCCFNet} formulates a multi-view neural network learning framework that combines MF-extracted features with user/item features. \cite{kanagawa2018cross} proposes a DA-based model, where stacked denoising autoencoders are used to extract item features, and a domain seperation network is responsible for recommendation. The above methods all rely on auxiliary contents (e.g. features of users, News, App) to learn shared patterns among domains while emphasize less on the patterns in the rating matrices. For instance, CCCFNet uses original MF method to represent users/items from rating matrices, which is inferior to state-of-the-art DL approaches. Recently, a work named collaborative cross networks (CoNet) \cite{hu2018conet} employs deep cross connection neural networks to learn and transfer shared rating patterns among domains without using any auxiliary contents.

Unlike all the previous works, our approach is different in four perspectives:\begin{inparaenum} \item we leverage DNN to learn linear and nonlinear shared rating patterns from rating matrices, which is more powerful than MF- or clustering-based methods; \item no dense data from the source domain is required for extracting transferable knowledge, as we rely on DNN to learn such knowledge directly from the raw sparse rating matrices; \item we do not need any auxiliary user/item features; \item we combine DA in our approach so that the possible distribution mismatch between two domains can be alleviated, which is more effective than simply using fully-connected neural networks to model the difference between two domains as CoNet \end{inparaenum}.

\section{Proposed Method}
\subsection{Problem Definition}
In this paper, we assume that the input data takes the form of explict feedbacks such as user ratings of items. We also assume that the source and target domains have the same set of users (i.e. the U-NI scenario), denoted by $\mathcal{U} = \{1,2,...,U\}$. The item sets from the source and target domains are $\mathcal{I^S} = \{1,2,...,I^S\}$ and $\mathcal{I^T} = \{1,2,...,I^T\}$ respectively, with the corresponding rating matrices given by $\mathcal{Y^S} = \{y_{ui}^S | u\in\mathcal{U}, i\in\mathcal{I^S}\}$ and $\mathcal{Y^T} = \{y_{ui}^T | u\in\mathcal{U}, i\in\mathcal{I^T}\}$. We denote the set of observed item ratings given by $u$ as $\mathcal{I}^S_u$ ($\mathcal{I}^T_u$), and the unobserved ones as $\mathcal{\bar{I}}^S_u$ ($\mathcal{\bar{I}}^T_u$) for source(target) domain. In each domain, the goal of recommendation can be specified as selecting a subset of items from $\mathcal{\bar{I}}^S_u$ ($\mathcal{\bar{I}}^T_u$) for user $u$ according to certain criteria that maximizes the user's satisfaction. In other words, the recommendation algorithms aim to give predictions on the unknown ratings of each user, i.e., $\hat{y}^S_{ui} (u\in\mathcal{U}, i\in\mathcal{\bar{I}}^S_u)$ or $\hat{y}^T_{ui} (u\in\mathcal{U}, i\in\mathcal{\bar{I}}^T_u)$. The values of $y^S_{ui}$ ($y^T_{ui}$) are numerical ratings in a certain range, e.g. $[1,5]$. The recommendation task we consider here is the single-domain problem \cite{khan2017cross}, where we predict $\hat{y}^T_{ui} (u\in\mathcal{U}, i\in\mathcal{\bar{I}}^T_u)$, via leveraging the information from the source rating matrix $\mathcal{Y^S}$. We use the widely-adopted metrics to measure the performance of all approaches used in this paper, such as RMSE defined by $\sqrt{1/(MN)\sum_{u=1}^{M}{\sum_{i=1}^{N}(\hat{y}^T_{ui} - y^T_{ui})^2}}$, where $M$,$N$ denote the number of users and items in the testing set respectively.

\subsection{Deep Domain Adaptation for CDR}
Figure~\ref{fig:DARec} shows the proposed model to integrate DA for shared-user CDR. The source and target domains share the same set of users, but the items are different. Thus, we first use AutoRec to learn a set of embeddings to represent each user's preferences. AutoRec \cite{sedhain2015autorec} leverages an autoencoder (AE) to predict the missing values in the rating matrix. U-AutoRec (I-AutoRec) first takes the partially observed rating vectors for each user (item), i.e. $\mathbf{y_{u}}$ ($\mathbf{y_{i}}$) as input, and maps each vector into a low-dimensional latent space, followed by a reconstruction layer as output to recover the rating vectors so that the original missing ratings can be generated for recommendation purpose. Formally, the reconstructed vectors are written as:
\begin{align} \label{eq:AutoRec:Pred}
    \mathbf{\hat{y}} = & h(\mathbf{W_2}\ g(\mathbf{W_1\ y+b_1})+\mathbf{b_2})
\end{align}
where $h(\cdot)$ and $g(\cdot)$ are activation functions, $\mathbf{W_1}$,$\mathbf{W_2}$, $\mathbf{b_1}$, $\mathbf{b_2}$ are corresponding weights and biases of the AE. The loss function is regularized square loss (U-AutoRec):
\begin{equation} \label{eq:AutoRec:Loss}
    \begin{aligned}
        loss_{AutoRec} = & \sum_{u=1}^{U}{\lVert \mathbf{\hat{y}_u} - \mathbf{y_u} \rVert}^2_\mathcal{O}\\
        +& \alpha ({\lVert \mathbf{W_1} \rVert}^2_{F}+{\lVert \mathbf{W_2} \rVert}^2_{F} + {\lVert \mathbf{b_1} \rVert}^2_2 + {\lVert \mathbf{b_2} \rVert}^2_2)
    \end{aligned}
\end{equation}
where ${\lVert \cdot \rVert}_F$ and ${\lVert \cdot \rVert}_2$ denote matrix Frobenius norm and vector $l^2$-norm respectively, $\mathcal{O}$ denotes the observed ratings in each vector $\mathbf{y_u}$, $\alpha$ controls the regularization strength.

After training the AutoRec, we calculate $\boldsymbol{f_u} = g(\mathbf{W_1}\ \mathbf{y_u}+\mathbf{b_1})$ with the trained parameters $\mathbf{W_1}$ and $\mathbf{b_1}$, as the low-dimensional latent factors for each user in both domains. $\boldsymbol{f_u}$ are then fed into a modified DANN network. To better extract the shared rating patterns from the latent space, the sequence of the latent factors for target domain are interleaved by those from the source domain.
To modify the original DANN \cite{ganin2015da} for rating prediction recommendation task, we use one deep feed-forward neural network with parameter set $\Theta_{r}$ to reconstruct the rating vectors for source and target domain seperately (rating predictor) and another neural network with parameter set $\Theta_{c}$ to predict the corresponding domain label $c\in\{0,1\}$, where $0$ denotes the input vector belongs to the source domain and $1$ the target domain (domain classifier). A fully-connected neural network is applied for feature extraction with parameter set $\Theta_{f}$ (rating pattern extractor). The training procedure to find the domain-invariant features involves optimizing on $\Theta_{f}$ to maximize the loss of the domain classifier, while optimizing on $\Theta_{c}$ and $\Theta_{r}$ to minimize the loss of the domain classifier and the label predictor. The total loss function is the combination of the predictor loss and the classifier loss:
\begin{equation} \label{eq:DANN:Loss}
    \begin{aligned}
        L(\Theta_{f},\Theta_{r},\Theta_{c}) =\ & loss_{pred}(\Theta_{f},\Theta_{r})- \mu\ loss_{dom}(\Theta_{c})
        + \lambda R
        \\=\ & \sum_{u=1}^{2U}{\lVert \mathbf{\hat{y}_{u,S}} - \mathbf{y_{u,S}} \rVert}^2_\mathcal{O}+\beta{\lVert \mathbf{\hat{y}_{u,T}} - \mathbf{y_{u,T}}\rVert}^2_\mathcal{O}
        \\+ \ & \mu \sum_{u=1}^{2U}{\hat{c}_u\log(c_u)+(1-\hat{c}_u)\log(1-c_u)}
        \\+ \ & \lambda R
    \end{aligned}
\end{equation}
where we employ binary cross-entropy for domain label classification loss ($\hat{c}_u$ is the predicted domain and $c_u$ is the ground-truth). $R={\lVert \Theta_{f} \rVert}^2 + {\lVert \Theta_{r} \rVert}^2 + {{\lVert \Theta_{c} \rVert}^2}$ is the regularizer term to alleviate overfitting, in which $\lambda$ controls the regularization strength. $\beta$ is used to balance the significance of the source and target parts in the total loss function. $\mu$ controls the portion of loss contributed by the domain classifier. Then the optimization objective can be expressed as:
\begin{equation} \label{eq:DANN:MinMax}
    \begin{aligned}
        (\hat{\Theta}_{f},\hat{\Theta}_{r}) = \  & \argmin_{\Theta_{f},\Theta_{r}}\ {L(\Theta_{f},\Theta_{r},\Theta_{c})} \\
        \hat{\Theta}_{c} = \  & \argmax_{\Theta_{c}} \ {L(\Theta_{f},\Theta_{r},\Theta_{c})}
    \end{aligned}
\end{equation}
To solve the above problem by stochastic grandient desent (SGD)-like algorithms, the original DANN introduces the gradient reversal layer (GRL). GRL is a simple structure that has different behaviours for forward and back propagation with no intrinsic parameters. In forward propagation, GRL acts as an identity function to let data pass through, while in backpropagation, it calculates the gradient from the subsequent level, multiplies it by $-\mu$ and then passes it to the preceding layer. Formally, GRL can be treated as a "function":
\begin{equation} \label{eq:DANN:GRL}
    \begin{aligned}
        &\Phi(\boldsymbol{x}) = \boldsymbol{x}\ &(forward\ propagation)\\
        &\frac{d\Phi(\boldsymbol{x})}{d\boldsymbol{x}} = -\mu\mathbf{I}\
        &(back\ propagation)
    \end{aligned}
\end{equation}
where $\mathbf{I}$ is the identity matrix. GRL is placed immediately after the rating pattern extractor and before the domain classifier. Therefore, the total loss function adapted to SGD-like algorithm is converted as:
\begin{equation} \label{eq:DANN:SGD_Loss}
    \begin{aligned}
    \tilde{L}(\Phi,\Theta_{f},\Theta_{r},\Theta_{c}) =\ &loss_{pred}(\Theta_{f},\Theta_{r})\\+\ &\ loss_{dom}(\Phi,\Theta_{c})
    + \lambda R
    \end{aligned}
\end{equation}
Equation \ref{eq:DANN:MinMax} is designed for extracting the deep shared rating patterns for each user in the target and source domain, which is called U-DARec. Our approach can also be applied on obtaining patterns from each item rating vector. We first use I-AutoRec to train a set of latent factors for all items in both domains, and feed them into the following DANN. As there is no overlap for items in the two domains, here, we extract the domain-exclusive rating patterns from the raw rating vectors and seperate them using the rating predictor. Thus, the loss of the domain classifier is minimized, leading to a different optimization objective from Equation \ref{eq:DANN:MinMax}:
\begin{equation} \label{eq:DANN:Min}
    \begin{aligned}
        (\hat{\Theta}_{f},\hat{\Theta}_{r},\hat{\Theta}_{c}) = \  & \argmin_{\Theta_{f},\Theta_{r},\Theta_{c}}\ {L(\Theta_{f},\Theta_{r},\Theta_{c})}
    \end{aligned}
\end{equation}
Consequently, the GRL layer is no longer needed and SGD-like algorithms can directly optimize Equation \ref{eq:DANN:Min}. We call this model I-DARec.

\renewcommand{\arraystretch}{1.2} 
\begin{table*}[!ht]
    \caption{Statistics of the Datasets}
    \centering
    \label{tab:datasets}
    \begin{tabular}{|c|c|c|c|c|c|c|c|c|c|}
        \hline
        \multicolumn{1}{|c|}{\bf{No.}} &
        \multicolumn{2}{c|}{\bf{Dataset}} &
        \multicolumn{2}{c|}{\bf{Item\#}} &
        \multicolumn{1}{c|}{\bf{User\#}} &
        \multicolumn{2}{c|}{\bf{Ratings\#}} &
        \multicolumn{2}{c|}{\bf{Sparsity}}\\
        \hline
        \hline
        & Source & Target & Source & Target & Shared & Source & Target & Source & Target\\
        \hline
        $1$ & Office Products & Movies\ and\ TV & $10,398$ & $21,732$ & $5,154$ & $40,294$ & $158,927$ & $99.92\%$ & $99.86\%$\\
        \hline
        $2$ & Sports\ and\ Outdoors & CDs\ and\ Vinyl & $16,420$ & $34,286$ & $5,713$ & $39,151$ & $79,019$ & $99.99\%$ & $99.96\%$\\
        \hline
        $3$ & Android\ Apps & Video\ Games & $9,185$ & $10,062$ & $2,034$ & $34,217$ & $21,312$ & $99.82\%$ & $99.90\%$\\
        \hline
        $4$ & Toys\ and\ Games & Automotive & $10,597$ & $7,375$ & $2,885$ & $25,103$ & $16,448$ & $99.92\%$ & $99.92\%$\\
        \hline
    \end{tabular}
\end{table*}

To train DARec, we apply mini-batch Adam algorithm for both user embedding training and rating prediction procedures. In AutoRec, we initialize $W_1, W_2$ with normal distribution (zero mean, $0.01$ standard deviation), $b_1, b_2$ with zeros and randomly draw a mini-batch $\mathcal{U}^{\prime}$ with size $B$ from $\mathcal{U}$ to obtain the embeddings. Then, the calculated $\boldsymbol{f_u}$ ($\boldsymbol{f_i}$ for I-DARec) in both domains are interleaved with each other, and are subsequently fed into the modified DANN. Similarly, we initialize the weights in DANN with normal distribution (zero mean, $0.01$ standard deviation), and biases with zeros. Then, we train the parameters $\hat{\Theta}_{f},\hat{\Theta}_{r},\hat{\Theta}_{c}$ in Equation \ref{eq:DANN:MinMax} or Equation \ref{eq:DANN:Min} for U-DARec and I-DARec respectively.

\section{Experiments}
In this section, we systematically evaluate the DARec model on multiple subsets extracted from the Amazon dataset with shared users in different categories. After giving the detailed dataset settings, we investigate the effectiveness of knowledge transfer in our approach and compare it with several state-of-the-art methods for cross-domain recommendation.
\subsection{Experimental Settings}
\subsubsection{Datasets}
We use the public dataset collected by J. McAuley \cite{he2016ups} and define different item categories as domains, where we select users with at least 5 ratings. We deliberately select categories that are as irrelavent as possible and the statistics are summarized in table~\ref{tab:datasets}. Both the source and target domains are extremely sparse with at least $99.8\%$ of the ratings are unobserved, which poses a major challenge on most clustering-based and MF-based CDR methods that normally require a relatively dense source domain.

\subsubsection{Parameter Settings}
In the embedding training stage, we leave out $10\%$ of the data as validation set to tune the hyperparameters where we adjust the number of hidden neurons from $100$ to $1500$ and regularizer coefficient from $0.1$ to $0.00001$ to get the lowest RMSE that can be achieved by AutoRec. Then, we feed all the data into the pre-trained AutoRec to get the embeddings calculated as $\boldsymbol{f_u}$ ($\boldsymbol{f_i}$). After the sequence of embeddings from source domain is interleaved with those from the target domain, we send it into the DANN. For the rating pattern extractor, we use only one layer with hidden neurons varying from $50$ to $500$. In the rating predcitor, we apply 3 layers for each domain and 2 layers for the domain classifier, where the number of neurons in each layer follows a pyramid shape and varies with the dimension of the previous pattern extractor. The parameter $\beta$ is changed from $0.0001$ to 1 so that the source domain is emphasized, and $\mu,\lambda$ are varied from $0.0001$ to $10,000$. To process the dataset, we randomly leave out $10\%$($20\%$) of the data for testing and $90\%$($80\%$) for training. $10\%$ of the training set is used as validation set for hyperparamater tuning.

\subsection{Effectiveness of Transfer Learning}

\begin{figure*}[!ht]
    \centering
    \includegraphics[width=\textwidth,height=3.8cm]{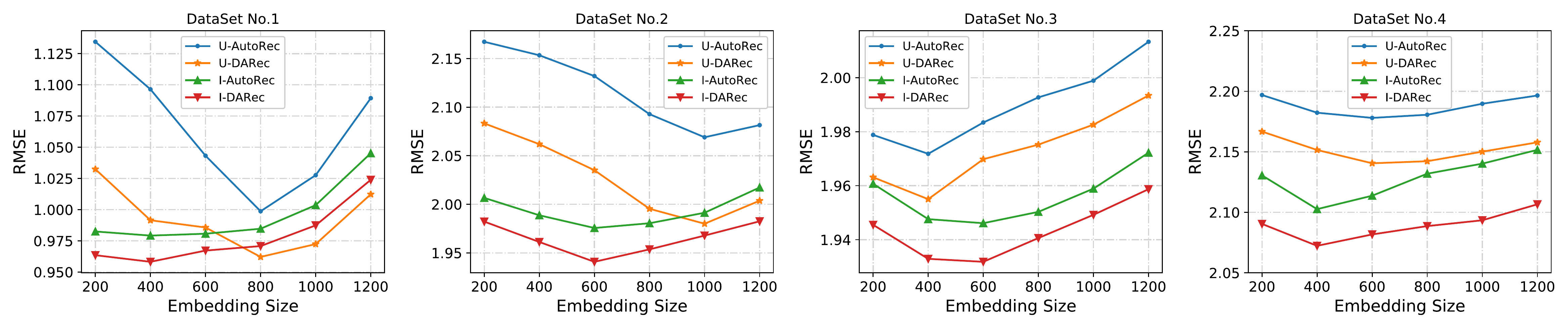}
    \caption{Impact of Embeddng Size on Prediction Accuracy}
    \label{fig:Embed}
    \vspace{-3mm}
\end{figure*}

We first investigate whether DARec can enhance the recommendation accuracy in the target domain. Apart from the standard AutoRec, we compare our approach with several state-of-the-art single-domain rating prediction models:
\begin{itemize}
    \item \textbf{PMF} \cite{mnih2008probabilistic} Probabilistic Matrix Factorization is one of the most populare MF methods. It obtains the predicted ratings from the inner product of corresponding user and target item latent factors.
    \item \textbf{RBM} \cite{salakhutdinov2007restricted} Restricted Boltzmann Machine uses neural networks to reconstruct the unknown ratings from the observed ones.
    \item \textbf{AutoRec} \cite{sedhain2015autorec} AutoRec is an RBM-like model which replaces the RBM with an AE to perform the unknown rating reconstruction procedure.
    \item \textbf{CF-NADE} \cite{zheng2016cfnade} Neural Autoregressive Distribution Estimator is applied for CF tasks where it replaces the role of RBM for rating reconstruction.
\end{itemize}
For PMF, we use the package Suprise\footnote{https://github.com/NicolasHug/Surprise}. For RBM, we rewrote a tensorflow version according to the opensource implementation\footnote{https://github.com/felipecruz/CFRBM} and for CF-NADE, we choose the version of I-CF-NADE-S and reworte a tensorflow version with reference to the opensource implementation\footnote{https://github.com/Ian09/CF-NADE}. We tune the latent factor dimensions, learning rate and regularizer coefficients to get their best performance.

\begin{figure}[ht]
    \includegraphics[width=0.48\textwidth]{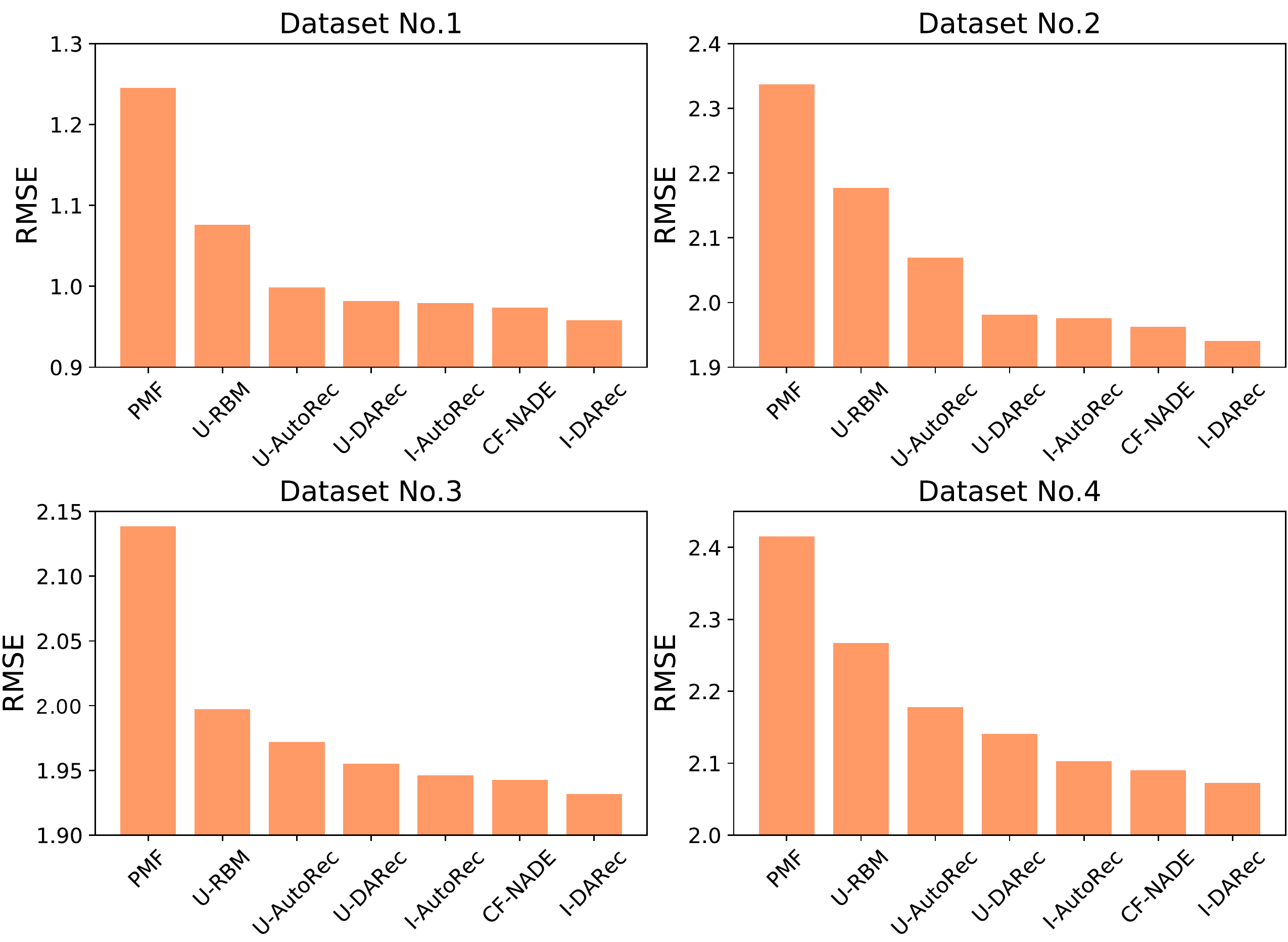}
    \caption{Comparison with Single-domain Approaches}
    \label{fig:effectiveness}
    \vspace{-3mm}
\end{figure}

Figure~\ref{fig:effectiveness} shows the average RMSE values of the compared single-domain recommendation methods, from which we can observe that both U-DARec and I-DARec obtain obvious improvment over U-AutoRec and I-AutoRec respectively. In U-DARec, the shared rating patterns are transferred via the DNN, while in I-DARec, the distinct rating patterns in each domain are extracted. For instance, in the Office Products and Movies \& TV dataset (No.1), U-DARec improves $3.66\%$ in RMSE compared with U-AutoRec and in the Sports \& Outdoors and CDs \& Vinyl dataset (No.2), the improvement is $4.30\%$. Although U-DARec still can not compete with I-AutoRec and CF-NADE, yet we achieve best performance by further improving on I-AutoRec via I-DARec. The superior performance from I-DARec is probably due to two reasons: \begin{inparaenum} \item the larger number of different input embeddings for I-DARec than that for U-DARec, making the training more effective for I-DARec; \item the adversarial objectives in U-DARec pose a trade-off between extracting shared patterns and enhancing rating prediction accuracy.\end{inparaenum}

\renewcommand{\arraystretch}{1.3} 

\begin{table*}[ht]
    \caption{Comparison with Baselines (\%: percentage of training data)}
    \label{tab:baselines}
    \centering
        \begin{tabular}{P{0.18\textwidth}P{0.085\textwidth}P{0.085\textwidth}P{0.085\textwidth}P{0.085\textwidth}P{0.085\textwidth}P{0.085\textwidth}P{0.085\textwidth}P{0.085\textwidth}}
        \hline
        \hline
        \multicolumn{1}{c|}{\textbf{Dataset}} &
        \multicolumn{2}{c|}{\thead{\textbf{Office Products} \\\textbf{\& Movies\ and\ TV}}} &
        \multicolumn{2}{c|}{\thead{\textbf{Sports\ and\ Outdoors} \\\textbf{\& CDs\ and\ Vinyl}}} &
        \multicolumn{2}{c|}{\thead{\textbf{Android\ Apps} \\\textbf{\& Video\ Games}}} &
        \multicolumn{2}{c}{\thead{\textbf{Toys\ and\ Games} \\\textbf{\& Automotive}}}\\
        \hline
        \hline
        \multicolumn{1}{c|}{\%} &
        \multicolumn{1}{c|}{$80$} &
        \multicolumn{1}{c|}{$90$} &
        \multicolumn{1}{c|}{$80$} &
        \multicolumn{1}{c|}{$90$} &
        \multicolumn{1}{c|}{$80$} &
        \multicolumn{1}{c|}{$90$} &
        \multicolumn{1}{c|}{$80$} &
        \multicolumn{1}{c}{$90$}\\
        \hline
        \multicolumn{1}{c|}{CMF} & $1.0825$ & $1.0583$ & $2.1043$ & $2.0578$ & $2.0784$ & $2.1295$ & $2.2636$ & $2.1855$ \\
        \hline
        \multicolumn{1}{c|}{CDTF} & $1.0577$ & $1.0389$ & $2.0702$ & $2.0126$ & $2.0461$ & $1.9965$ & $2.2234$ & $2.1680$ \\
        \hline
        \multicolumn{1}{c|}{FM-CDCF} & $1.0272$ & $0.9925$ & $2.0580$ & $1.9942$ & $1.9963$ & $1.9722$ & $2.1908$ & $2.1534$ \\
        \hline
        \multicolumn{1}{c|}{CoNet} & $0.9782$ & $0.9689$ & $1.9835$ & $1.9587$ & $1.9744$ & $1.9562$ & $2.1507$ & $2.1356$ \\
        \hline
        \hline
        \multicolumn{1}{c|}{U-DARec} & $0.9985$ & $0.9821$ & $2.0123$ & $1.9808$ & $1.9876$ & $1.9550$ & $2.1734$ & $2.1405$ \\
        \hline
        \multicolumn{1}{c|}{\bf{I-DARec}} & $\mathbf{0.9665}$ & $\mathbf{0.9582}$ & $\mathbf{1.9673}$ & $\mathbf{1.9408}$ & $\mathbf{1.9546}$ & $\mathbf{1.9318}$ & $\mathbf{2.1152}$ & $\mathbf{2.0723}$ \\
        \hline
        \hline
        \end{tabular}
    \vspace{-3mm}
\end{table*}

We then investigate the impact of the embedding size trained by AutoRec on the single-domain recommendation performance. We first fix the structure of the DANN(i.e., the number of hidden neurons of rating pattern extractor is set to 100) and then change the dimension of the embeddings from 200 to 1200 for all datasets. From figure~\ref{fig:Embed}, it can be observed that for both U-DARec and I-DARec, there exist optimal embedding sizes. For instance, in the Office Products and Movies \& TV dataset (No.1), U-DARec reaches minimum RMSE at around $800$ and I-DARec at around $400$. When the input embedding dimension is not large enough, the rating patterns can not be effectively extracted, while at a large embedding size, the learned embeddings are too noisy with undesired latent feature details, bringing detrimental effects on the overall performance. Furthermore, we also notice that although I-DARec achieves the lowest RMSE in general, yet the enhancement from U-DARec is greater than that from I-DARec, which can be well understood from the aligned-user settings of the datasets as the model is more effective in learning the shared rating patterns.

\subsection{Comparison with Baselines}
We further compare our DARec model with some of the following cross-domain recommendation approaches:
\begin{itemize}
    \item \textbf{CMF} \cite{singh2008CMF} Collective Matrix Factorization (CMF) is a latent factor model that establishes relationships among multiple domains via MF. Our experiment settings correspond to the three-entity-type schema with two domains. The first entity type is the items in the source domain and the third entity type is the items in the target domain, with the shared users as the bridging entity type. A set of latent factors is learned for each entity type and rating predictions are made via inner products of corresponding latent factors. It is a MF-based transfer learning approach that jointly factorizes two related domains.
    \item \textbf{CDTF} \cite{hu2013personalized} Cross Domain Triadic Factorization (CDTF) is another latent factor model that leverages the triadic relation user-item-domain. It learns the triadic factors for user, item and domain via factorizing a three-order tensor with weighted square loss, where each slice of the tensor corresponds to one domain. CDTF matches the CDR scenario we consider, where each domain has equivalent number of users but vary in the number of items.
    \item \textbf{FM-CDCF} \cite{Loni2014cross} Factorization Machine Cross-Domain Collaborative Filtering (FM-CDCF) is based on FM, where the feature vector in single-domain FM is extended to incorporate collaborative information from other domains. It is well suited for the aligned-user CDR scenario where each input feature vector is composed of the one-hot representation of the user $u$, one-hot representation of the item $i$ in the target domain, and all the normalized ratings given by $u$ in the source domain. It is still a shallow model without the ability of capturing nonlinear interactions between domains.
    \item \textbf{CoNet} \cite{hu2018conet} Collaborative Cross Networks (CoNet) is a state-of-the-art deep CDR model that transfers knowledge between domains with aligned users. It uses DNN to learn latent features in each domain and applies a modified cross-stitch neural network to enable knowledge transfer between two adjacent layers. It is a highly competitive method and achieves best performance among other non-DL approaches.
\end{itemize}
For CMF, we rewrote a python version based on the original Matlab code\footnote{http://www.cs.cmu.edu/~ajit/cmf/}. For FM-CDCF, we adopt the libfm implementation\footnote{http://www.libfm.org/} for FM and feed in the extended input feature vectors. For CDTF and CoNet, we implement our own python version according to the original papers. For all the baselines, we tune the latent factor dimensions, learning rate and regularizer coefficients to get their best performance.

Table~\ref{tab:baselines} reports the comparison results among the selected baselines for CDR. Note that in each domain pair, the chosen categories are as irrelevant as possible, so that we can better examine the performance of our model on learning and transfering knowledge from the rating matrix only. It is obvious that DL-based approaches (i.e., CoNet and DARec) are superior to non-deep models. This can be explained by the ability of nonlinearity modelling from the deep neural networks. Also, MF-based mehtods perform better when the source domain dataset is relatively dense, but the datasets we consider in this work are of high sparsity (over $99.8\%$). For example, in the Office Products and Movies \& TV dataset with $90\%$ training data, CoNet gains $2.38\%$ RMSE improvement over FM-CDCF and I-DARec achieves $3.45\%$ over FM-CDCF. This result shows that DL-based methods are more suitable for extreme sparse datasets, which is in consistency with the findings in \cite{hu2018conet}. Furthermore, we notice that CoNet performs better than U-DARec as U-DARec adopts adversarial training to balance the extraction of shared user rating patterns and the pursuit of high rating prediction accuracy. This weakness is overcome by I-DARec which minimizes the losses of domain classifier and rating predictor at the same time. Thus, as we can see, I-DARec achieves the best performance among the baselines.

\section{Conclusions}
In this paper, we proposed a deep domain adaptation model for cross-domain recommendation, where we consider the single-domain recommendation task with users aligned. Different from most of the previous works that utilize contextual information to complete the knowledge transfer, we only rely on the rating matrices. To extract the shared user rating patterns from the two domains, we first apply AutoRec to generate abstract embeddings to represent each user. Then, the embeddings from the source and target domains are interleaved and fed into a deep domain adaptation neural network to complete the transfer process. Using a domain classifier, the shared deep rating patterns from the two domains are extracted, followed by a rating predictor that seperates the estimated ratings for each domain. Apart from leveraging shared features for each user, we extend our model through taking the ratings for each item as input, where we extract the distinct features of the two items in different domains. By comparing with several single-domain recommendation methods, we demonstrate the ability of transfering knowledge of our proposed model. Furthermore, we demonstrate that our model achieves best performance in rating prediction task among several widely-adopted and state-of-the-art cross-domain recommendation approaches. In the future, we will integrate content-based domain adaptation models to gain further performance improvement.
\newpage
\bibliographystyle{named}
\bibliography{ijcai19}

\end{document}